\documentclass[journal,12pt,onecolumn,final,]{IEEEtran}
\IEEEoverridecommandlockouts
\usepackage{cite}
\usepackage{amsmath,amssymb,amsfonts}
\usepackage{graphicx}
\usepackage{caption}
\usepackage{subcaption}
\usepackage{textcomp}
\usepackage{url}
\usepackage{xcolor}
\usepackage{color, colortbl}
\definecolor{Gray}{gray}{0.9}
\definecolor{LightGray}{gray}{0.95}

\def\BibTeX{{\rm B\kern-.05em{\sc i\kern-.025em b}\kern-.08em
    T\kern-.1667em\lower.7ex\hbox{E}\kern-.125emX}}
\begin{document}

\title{A Structured Analysis of the Video Degradation Effects on the Performance of a Machine Learning-enabled Pedestrian Detector\thanks{The research has been supported by Chalmers AI Research Centre (CHAIR).}
}

\author{\IEEEauthorblockN{Christian Berger}\\
\IEEEauthorblockA{\textit{University of Gothenburg, Sweden} \\
\textit{Department of Computer Science \& Engineering}\\
christian.berger@gu.se}
}

\maketitle

\begin{abstract}
Machine Learning (ML)-enabled software systems have been incorporated in many
public demonstrations for automated driving (AD) systems. Such solutions have
also been considered as a crucial approach to aim at SAE Level 5 systems, where
the passengers in such vehicles do not have to interact with the system at all
anymore. Already in 2016, Nvidia demonstrated a complete end-to-end approach
for training the complete software stack covering perception, planning and
decision making, and the actual vehicle control. While such approaches show
the great potential of such ML-enabled systems, there have also been demonstrations
where already changes to single pixels in a video frame can potentially lead to
completely different decisions with dangerous consequences in the worst case.
In this paper, a structured analysis has been conducted to explore video
degradation effects on the performance of an ML-enabled pedestrian detector.
Firstly, a baseline of applying ``You only look once'' (YOLO) to 1,026 frames
with pedestrian annotations in the KITTI Vision Benchmark Suite has been
established. Next, video degradation candidates for each of these frames
were generated using the leading video compression codecs libx264, libx265,
Nvidia HEVC, and AV1: 52 frames for the various compression presets for color
frames, and 52 frames for gray-scale frames resulting in 104 degradation candidates
per original KITTI frame and in 426,816 images in total. YOLO was applied to
each image to compute the intersection-over-union (IoU) metric to compare the
performance with the original baseline. While aggressively lossy compression settings
result in significant performance drops as expected, it was also observed that some
configurations actually result in slightly better IoU results compared to the
baseline. Hence, while related work in literature demonstrated the potentially
negative consequences of even simple modifications to video data when using
ML-enabled systems, the findings from this work show that carefully chosen
lossy video configurations preserve a decent performance of particular ML-enabled
systems while allowing for substantial savings when storing or transmitting data.
Such aspects are of crucial importance when, for example, video data needs to be
collected from multiple vehicles wirelessly, where lossy video codecs are required
to cope with bandwidth limitations for example.
\end{abstract}

\section{Introduction}
Automated driving (AD) systems are said to positively contribute to further lower
fatally injured traffic participants like vulnerable road users such as pedestrians
or bicyclists. Such systems continuously monitor a vehicle's surroundings with
multiple sensors such as cameras, radars, and laser scanners to get a reliable
representation of the objects around a vehicle to plan safe driving trajectories
or to conduct evasive maneuvers to avoid collisions. However, as tragically reported
by the first fatal accident of a pedestrian caused by a self-driving
car\footnote{The system was not classified as SAE 4 or SAE 5 Level system as a safety driver was required for its operation.}
involving an Uber test vehicle in 2018, the current development of such systems is
still far away from deploying safe solutions that work reliably in harsh and
challenging contexts to achieve SAE 4 Level systems; ie., AD systems that for
certain operational design domains (ODD) allow the driver to completely hand over
the driving tasks to the machine and not acting as a fallback solution to the
system if something unexpected is happening.

\subsection{Motivation}
Major substantial breakthroughs when using ML-enabled systems for AD have been
demonstrated; a remarkable example was presented in 2018 by Bojarski et al.~from
Nvidia (cf.~\cite{nvidia}) who showcased a neural network (NN) that covered the complete
end-to-end process from perceiving a video frame to a resulting driving decision
as a steering wheel angle by just using 72 hours of video data for training.
This demonstration was a remarkable cornerstone in the global competition of
developing AD systems. Nevertheless, countless edge cases to safely handle like
partially occluded pedestrians carrying unexpected items that are not captured
in large scale sensor recordings need to be identified, labeled, and appropriately
used for training and validation of ML-enabled systems.

To capture such data, hundreds of test vehicles with a broad range of sensors are
needed to increase the likelihood of facing such unexpected edge cases. Tesla, as
already stated in their 2015 financial report, provided an ``\dots advanced set of
hardware including a forward radar, a forward-looking camera, 12 long-range ultrasonic
sensors, and a high-precision digitally controlled electric assist braking system''
in their vehicles; and, more importantly, this ``hardware suite, along with
over-the-air firmware updates and \emph{field data feedback loops} from the onboard
camera, radar, ultrasonics, and GPS, enables the system to continually learn and
improve its performance.''\footnote{Cf.~\url{https://t1p.de/tesla-10k-2015}}

\subsection{Problem Domain and Motivation}
To realize such field data feedback loops, data needs to be transferred wirelessly
from customer vehicles for further analysis at the automotive OEM. However, wireless
bandwidth is limited and hence, simply transferring raw data from all sensors is
practically and economically infeasible. Therefore, smart choices have to be made
regarding what scenarios to capture and transfer, and how to compress such data in
a potentially lossy way while discarding as little valuable information as possible.

\subsection{Research Goal and Research Questions}
The goal for this work is to conduct a structured analysis to understand the potential
impact of degraded video on the performance of an ML-enabled system. The following
research questions have been derived:

\begin{description}
\item[RQ-1] What is the baseline performance of a state-of-the-art ML-enabled object
detector (YOLO) when applied to detect pedestrians in video frames from the KITTI Vision
Benchmark Suite?
\item[RQ-2] How does the performance of the ML-enabled object detector change when the
video frames are systematically degraded in quality using state-of-the-art lossy video
encoders?
\end{description}

\subsection{Contributions}
This work contributes with a structured analysis of the impact of degraded video on
the performance of a state-of-the-art ML-enabled object detector. The study includes
the following leading state-of-the-art video compression algorithms: h264 as provided
by libx264, h265 as provided by libx265, HEVC provided in the hardware-accelerated
implementation as found in Nvidia GPUs, and the upcoming AV1 codec on a broad set of
examples from a widely adopted and accepted dataset.

\subsection{Scope and Delineations}
The left color images of the object dataset from the KITTI Vision Benchmark Suite (cf.~ Geiger et al.~\cite{GLU12})
were used as dataset for this study. The video compression is conducted using the video
encoder's provided compression levels for both, colored and gray-scale variants. For
each generated candidate image, the the peak signal-to-noise ratio (PSNR) has been
determined to quantify the image quality degradation.

\subsection{Structure of the Article}
The remainder of the article is structured as follows: In Sec.~\ref{sec:related-work},
relevant related work is outlined and discussed. Sec.~\ref{sec:methdology} describes
the experimental setup followed by the presentation of the results in Sec.~\ref{sec:results}.
In Sec.~\ref{sec:analysis-discussion}, the findings are analyzed and discussed before
the paper concludes and provides an outlook for future work in Sec.~\ref{sec:conclusions}.

\section{Related Work}
\label{sec:related-work}

The amount of data from a prototypical self-driving car as of today can easily exceed
0.75GB/s\footnote{Cf.~\url{https://t1p.de/750MBperSec}} due to high-resolution cameras,
laser scanners, and radars. As outlined by our systematic literature survey, though,
video streams in datasets used for training ML are typically compressed using lossy
encoders to reduce the data size (cf.~\cite{KYB19}). However, Koziarski and Cyganek
showed that contemporary architectures for neural networks (NNs) are significantly
affected by reduced image quality (cf.~\cite{KC18}); similarly, Hosseini et al.~showed
that even professionally trained NNs like Google’s Cloud Vision API are easily misled
when adding perturbations to video feeds (cf.~\cite{HXP17}).

Ongoing research to increase the robustness of ML-based approaches is primarily focusing
on using a high-quality input dataset during the training of ML-systems but improving
its performance when presenting input samples of lower quality afterwards. Publications
herein typically study ML- based approaches that return a label for a given input data
sample; contemporary publications extend or improve the NN ``VGG16'' for this task originally
presented by Simonyan and Zisserman (cf.~\cite{SZ14}).

Results as presented by Koziarski and Cyganek (cf.~\cite{KC18}) observe that contemporary
architectures for NN are sensitive to degraded quality of input samples. They note that even
mild changes to an input sample, which are hardly observable by a human eye, often result in
a noticeable drop in an NN's performance. The authors also showed this effect across different
NN architectures.

Binachi et al.~(cf.~\cite{BVPB19}) also confirm this observation as today's NN fail ``to
generalize whenever they are presented different conditions from the ones that they have
encountered during training.'' They suggest applying fine-tuning those layers of a NN that
are sensitive to quality degradation. Sun et al.~(cf.~\cite{SOZ+18}) propose feature
quantization to improve the robustness of a NN handling distortion effects more reliably.
For NN architectures embodying their suggested additional non-linear layer, the performance increased regarding generalization and various image distortions.

Roy et al.~(cf.~\cite{RGBP18}) also confirm the aforementioned findings. In addition, they
report that reducing the quality of an input sample even further unexpectedly resulted in
a better performance compared to high quality samples, which seems intuitively contradictory
(cf.~\cite{RGBP18}, Fig.~5(f)).

Ghosh et al.~(cf.~\cite{GSA+18}) studied effects of JPEG image compression on the performance
of convolutional NNs (CNNs) and they observed a ``dependency on the relative sizes of the
compression units [\dots] and the receptive field size of the CNN being used for
classification.'' Hence, this dependency strongly encourages the research direction outlined
in this proposal taking dependencies of lossy data compression approaches on a ML's architecture
explicitly into account. Usama and Chang (cf.~\cite{UC18}) agree with the aforementioned
authors' observations and also confirm that fine-tuning the affected layers within a NN is
``more practical than re-training.''

All aforementioned recent works agree in unison that today's NNs are affected by quality
degradation. However, to the best knowledge of the author, this is one of the first studies
that systematically analyzes the effects from stream-oriented compression artifacts as
originating from, for example, wide-spread video codecs on the performance of a NN.

\section{Methodology}
\label{sec:methdology}

\begin{figure*}[th!]
 \centering
 \includegraphics[width=\textwidth]{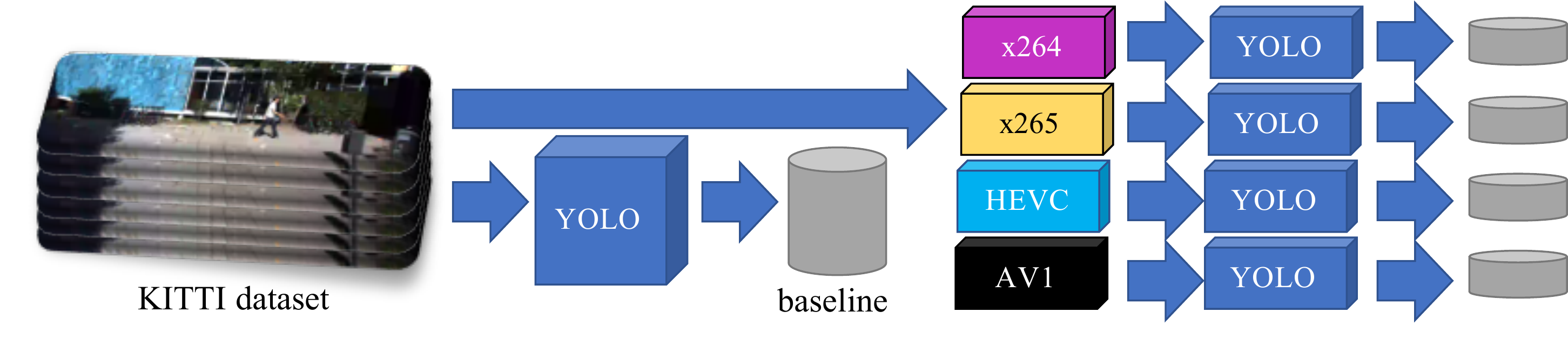}
 \caption{The experimental setup to analyze the effects from applying several leading video compression codecs to the KITTI dataset (cf.~ Geiger et al.~\cite{GLU12}): The original dataset was used to select all images that contain pedestrian labels (1,026 frames) and the YOLO object detector was applied to defined a baseline performance shown as blue in subsequent charts. Next, the dataset was used and the following four video compression codecs with their constant rate factor-based encoding configurations (CRF) ranging from CRF-0 meaning best possible quality, to CRF-51 meaning lowest quality, were applied to each of the 1,026 frames to create a set of degraded images, to which the YOLO detector was applied again to obtain the data from the performance impact. The following color coding is used for this paper: The h264 codec based on libx264 is depicted in the following charts using magenta, the h265 codec based on libx265 is shown as yellow, HEVC using Nvidia's hardware acceleration as found in their GPUs as dedicated silicon implementation is depicted in cyan, and the upcoming video codec AV1 from the Alliance for Open Media is shown in black.}
 \label{fig:ExperimentalSetup}
\end{figure*}

The overall experimental setup for our research goal and to address our research questions
is depicted in Fig.~\ref{fig:ExperimentalSetup} and described in the following. We are using
the KITTI Object Detection Evaluation dataset (left camera)\footnote{Cf.~\url{https://t1p.de/kitti}}
with their annotations in our data analysis pipeline. The original KITTI labels have been
converted to the YOLO annotation format using an open source Python
tool\footnote{Cf.~\url{https://github.com/ssaru/convert2Yolo}}. After the labels have been
transformed, a subset of the data was created to only choose those images that contain a
single pedestrian. We focus on a single pedestrian per frame to allow for a systematic
analysis on the impact on this specific category as they are the most vulnerable road users;
hence, the performance of any ML-enabled perception system needs to be analyzed in particular
to achieve a reliable detection of such traffic participants. The resulting subset of images
contained 1,026 images.

Next, the YOLOv3 object detector (cf.~Redmon et al.~\cite{RDGF16}) based on a Python
implementation\footnote{Cf.~\url{https://github.com/shahkaran76/yolo_v3-tensorflow-ipynb}}
was applied to obtain predictions about where within the original images the pedestrians were
located to establish a baseline for further comparisons. In the result charts presented in
Sec.~\ref{sec:results}, the baseline performance is denoted using the color blue. The
thresholds for Intersection-over-Unions (IoUs) and confidence for accepting a detection are
set to 0.5, respectively.

Next, we applied different video compression codecs to the 1,026 images in our dataset.
We chose the following four codecs: (1) The h264 video codec as realized in the libx264
implementation\footnote{Cf.~\url{https://www.videolan.org/developers/x264.html}} that provides
the ``best-in-class performance and compression'', (2) the h265 video codec as realized
in the libx265 implementation\footnote{Cf.~\url{https://www.videolan.org/developers/x265.html}},
(3) the hardware-accelerated High Efficiency Video Coding (HEVC) as realized in Nvidia
GPUs, and (4) the upcoming AV1 video codec provided by the Alliance for Open
Media\footnote{Cf.~\url{http://aomedia.org/av1}}. The versions of the video codecs in use
for the experimental setup are shown in Tab.~\ref{tab:versions}.

\begin{table}[h!]
\centering
\begin{tabular}{|p{2cm}|p{5cm}|}
\hline
\textbf{Video Codec} & \textbf{Version} \\
\hline
\hline
\rowcolor{LightGray}
libx264 & \texttt{152\_0.152.2854+gite9a5903-2} \\
libx265 & \texttt{146\_2.6-3} \\
\rowcolor{LightGray}
HEVC & \texttt{Nvidia drivers 460.39} \\
\rowcolor{LightGray}
     & \texttt{CUDA 11.2} \\
\rowcolor{LightGray}
     & \texttt{GPU GeForce RTX 2070} \\
AV1 & \texttt{2.0.2} \\
\hline
\end{tabular}\vspace{0.1cm}
\caption{The versions of the video codec in use.}
\label{tab:versions}
\end{table}

\begin{figure*}[thb]
     \centering
     \begin{subfigure}[b]{0.45\textwidth}
         \centering
         \includegraphics[width=\textwidth]{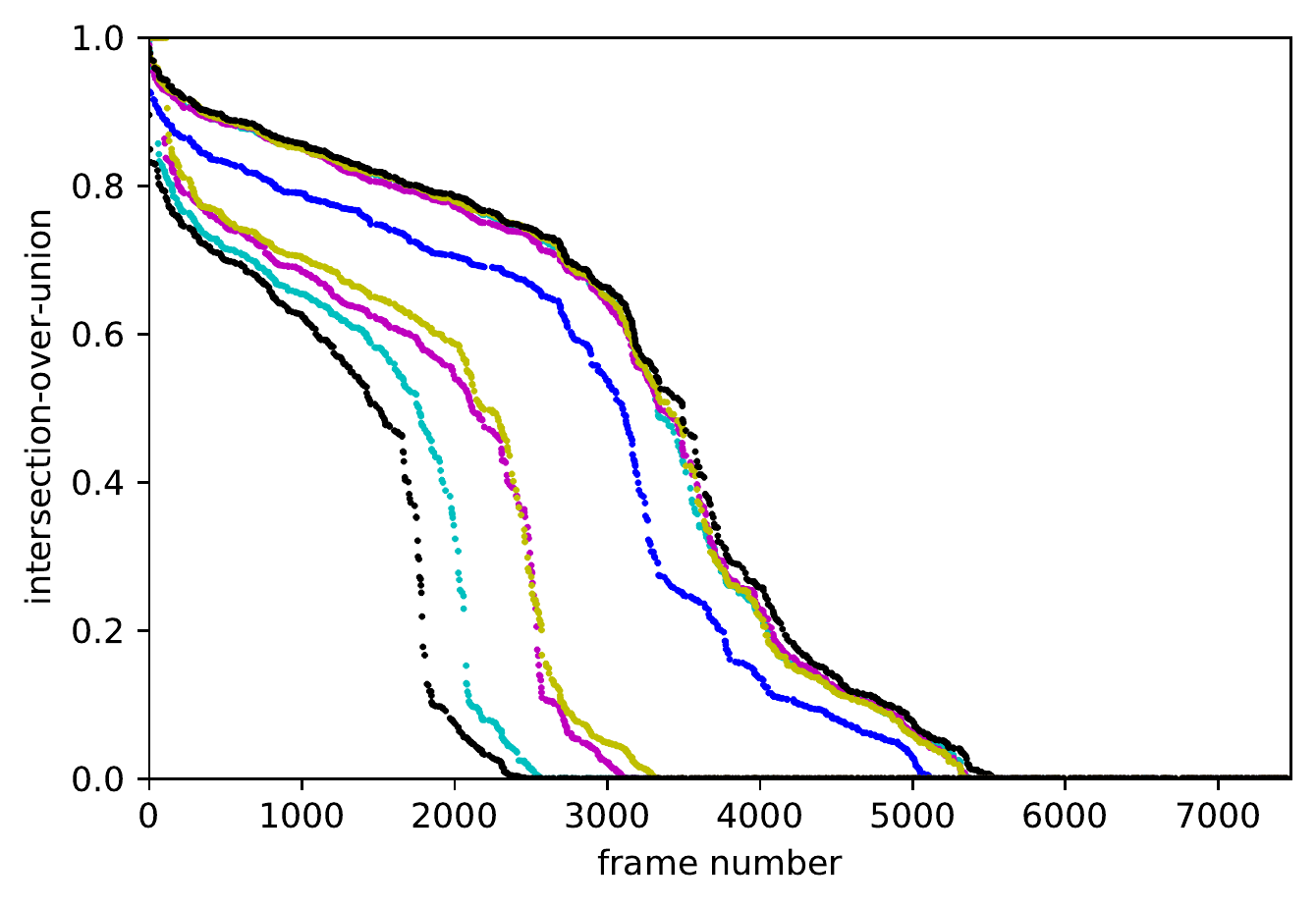}
         \caption{Performance of YOLO on colored input frames.}
         \label{fig:YOLO-Performance-Color}
     \end{subfigure}
     \hfill
     \begin{subfigure}[b]{0.45\textwidth}
         \centering
         \includegraphics[width=\textwidth]{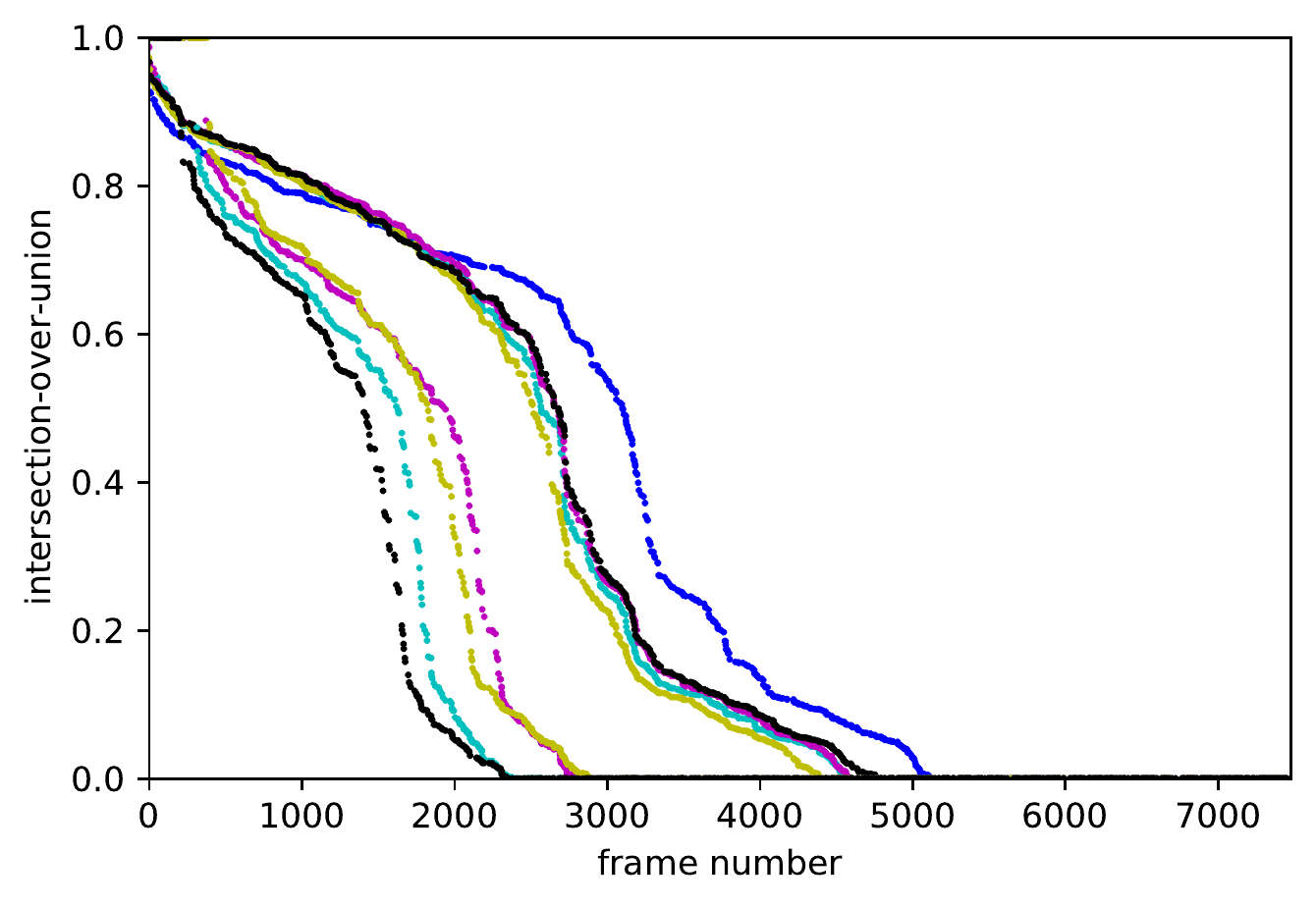}
         \caption{Performance of YOLO on gray-scale input frames\footnotemark}
         \label{fig:YOLO-Performance-Gray}
     \end{subfigure}
     \caption{Performance of the YOLO object detector applied to the subset of KITTI images containing only a single pedestrian: The x-axis represents the frame number from the KITTI dataset and the y-axis the Intersection-over-Union (IoU) value as calculated between the ground truth label and the prediction. In blue, the performance of YOLO on the unmodified dataset is shown. Similarly to Fig.~\ref{fig:ExperimentalSetup}, the h264 codec is depicted by using magenta, the h265 codec is shown as yellow, HEVC using the Nvidia GPU is represented as cyan, and the video codec AV1 is shown in black.}
     \label{fig:YOLO-Performance}
\end{figure*}

For each video codec, we selected the constant rate factor (CRF)-based encoding
configurations for a constant quality. The codecs provide the quality range from 0
meaning best possible quality, to 51 meaning lowest quality but smallest file size on disk.
Furthermore, gray-scale variants for each of the candidate frames created by the
respective video codecs were created extending the number of images for our
experiments to 1,026 original frames, multiplied by four video codecs with 52 CRF
settings each for two color spaces (RGB and gray-scale) resulting in 426,816 candidate
images. To avoid changing the internal architecture of the NN, the color images were
transformed from RGB to gray-scale (ie., reducing the color channels to one) and
transforming them back to RGB (ie., recreating all three color-channels containing
only the gray-scale information in all of them).

\section{Results}
\label{sec:results}

For each of the 1,026 images from the KITTI, we applied the four video codecs for
each of the 52 encoding settings for CRF-based compression. Furthermore, we also
converted the resulting degraded images to gray-scale. Hence, the resulting in
426,816 frames in total.

Finally, the YOLOv3 detector was applied to all these frames to obtain the prediction
results, consisting of location of a bounding box of detected pedestrian as well as
the detection confidence.

\footnotetext{The baseline performance is computed on the unmodified input images.}

In this section, we are presenting the results from our experiments to address
the research questions. The first RQ is focusing on determining the baseline
performance of a state-of-the-art ML-enabled object detector when applied to detect
pedestrians in video frames. As shown in Fig.~\ref{fig:YOLO-Performance-Color}, the
performance of the object prediction are shown in the selected subset of 1,026
images from the KITTI dataset. The performance of the object detector YOLO on the
unmodified KITTI dataset is plotted using the color blue. The results are sorted in
descending order; on the x-axis, the frame number from the KITTI dataset is shown and
on the y-axis, the Intersection-over-Union (IoU) value that can range from 1.0
representing a perfect match between prediction and annotation (ie., the ideal case),
and 0 meaning no match (ie., not detected or false positive at other location).
The baseline performance is greater than 0.5 (ie., the prediction bounding box is
having an overlay with the ground truth label of at least 50\%) in approximately half
of the cases. The baseline performance also drops significantly from approximately
0.6 to 0.25 during only a small range of images (frame identifiers between 3,000 and
4,000) before the lowest performance is achieved between 0.2 and 0 between frame
identifiers 3,500 and approximately 5,000.

The second RQ is investigating how the performance of the ML-enabled object detector
changes when the video frames are systematically degraded in quality using
state-of-the-art lossy video encoders. The same chart (cf.~Fig.~\ref{fig:YOLO-Performance-Color})
also depicts the performance of the four video codecs: The h264 video codec is represented
in magenta, the h265 video codec is plotted as yellow, the hardware-accelerated HEVC
video codec implemented by Nvidia GeForce RTX 2070 is depicted in cyan, and the AV1
video codec is shown in black. Furthermore, for each codec we have plotted the best
performing constant rate factor setting above the blue line, and their worst
performance from the most lossy CRF configuration profile below the blue line.

It is surprising that applying a video codec for lossily compressing the video data
is resulting in an \emph{improvement} of the detection rate: The baseline performance
of YOLO is matching approximately 68.8\% of the original ground truth labels. After
applying the h264 video codec, the performance increases to 72.2\%; h265 matches
72.1\% and the hardware-accelerated HEVC codec matches 72.4\% of the ground truth
labels. The upcoming AV1 codec is able to match 73.6\% of the ground truth labels.
Hence, not all ground truth labels are found with the pre-trained YOLO implementation.

Fig.~\ref{fig:YOLO-Performance-Gray} shows the performance of the YOLO object detector
when fed with gray-scale data. The performance drops in comparison with the color input;
however, the performance drop is much more significant for the yet best performing
CRF configurations that now also drop largely below the baseline performance. However,
for the best performing CRF configurations on images with high confidence levels, the
color information does not seem to contribute significantly to the prediction performance
as the video compressed candidate images are still on or above the baseline performance.
For the gray-scale case, the performance from h264 drops to 61.6\%; h265 matches
only 58.8\% and the hardware-accelerated HEVC codec matches 61.5\% of the ground truth
labels. The AV1 codec is manages to match 63.5\% of the ground truth labels and thus,
also performing best from all four video codecs.

\section{Analysis and Discussion}
\label{sec:analysis-discussion}

The goal for this study to determine the baseline performance of a state-of-the-art
object detector when applying it to images containing pedestrians following by an
analysis how this performance is affected when applying lossy video codecs. As shown
in Fig.~\ref{fig:YOLO-Performance-Color}, the object detector is working reasonably
well when applying it in an unmodified variant directly on the input dataset. While
its performance is arguably not sufficient for a potential usage in production
systems with a performance of only 68.8\%, it served well the purpose of setting 
up an experiment for analyzing the effect from lossy video codecs.

Unexpectedly, the performance of the object detector did increase \emph{after}
applying a lossy video codec to the input data. One the one hand, this observation
is confirming the findings as reported by Roy et al.~(cf.~\cite{RGBP18}) and on the
other hand, it may raise the question how possible software architectures for
ML-enabled systems would actually \emph{benefit} from applying some lossy video
compression first before feeding a NN with input data for obtaining predictions.
However, while this unintuitive observation may lead to this conclusion, the actual
compression settings may vary from situation to situation (eg., because of the frame's 
content or light conditions) and hence, a generalization cannot be derived from this
observation.

Furthermore, applying a lossy video codec is irrecoverably discarding potentially
relevant information from the environment that may be of interest for future
changes of the NN or when re-training a NN after changing its structure. Hence,
while applying lossy video codecs to save disk space is tempting, preserving as much
(or preferably all) information from the sensors during a data collection is nevertheless
recommended and applying lossy video codecs shall only be applied either during
pos-processing of the data (ie., when the effects of information loss can be analyzed
systematically as described here) or for archival purposes.

As possible threats to validity, we can list that we did neither specifically re-train
the object detector on the input data to improve the overall baseline performance,
nor did we re-train the detector to better cope with the gray-scale data. However,
as the NN was used in an unmodified and un-optimized way for all experiments, this
possible threat was mitigated. Furthermore, we did not look specifically in sequences
of frames from the same situation covering a longer duration where the detection
performance may even vary for the same scenario.

\section{Conclusions and Future Work}
\label{sec:conclusions}

In our study, we present the results from applying a state-of-the-art ML-enabled
object detecting algorithm to detect pedestrian in a broadly accepted dataset
(KITTI vision benchmark suite). After having established this baseline performance,
we systematically degraded the quality of the input images by applying four
video codecs (h264, h265, hardware-accelerated HEVC, and AV1) to the dataset covering
all their possible constant rate factor settings. While the performance of the
object detector did drop significantly for the lowest quality as expected, we
observed an increase in detection performance \emph{after} having applying the
lossy video codec to the dataset. Our study shows that the upcoming AV1 codec is
showing the best results followed by the hardware-accelerated HEVC codec using
an Nvidia GeForce RTX 2070. However, the current software implementation of the
AV1 codec is performing quite slow currently not allowing a potential real-time
application, while the HEVC encoder shows a real-time-capable performance due to
its hardware-acceleration. However, in contrast to HEVC, AV1's aim is to provide
an open and royalty-free video codec technology stack, which is of interest of
content providers as well as companies creating products using this technology.
For instance, for self-driving vehicles for example, discussions are currently
ongoing regarding event-data recorders (EDR) that shall continuously record data
from the environment to be used for post-mortem analysis in case of accidents for
example. In this case, the trade off between best possible quality of data to
analyze and understand the (mis-)behavior of a NN is needed; however, the still
growing amount of data to be handled by a self-driving vehicle may require the
compromise of using lossy video codecs to store as much data as possible yet in
a decent quality useful for analysis.

Future extensions to this work may investigate the impact of sequences of images
of the same scenario on the effects of applying lossy video codecs to input data
fed to NN. Furthermore, the potential of NN-enabled video compression may be
investigated to compare that direction with the traditional video compression
strategies.

\section*{Acknowledgments}

The research presented in this work has been supported by Chalmers AI Research
Centre (CHAIR) for the project DegradeFX.

\bibliographystyle{IEEEtran}
\bibliography{literature}

\begin{thebibliography}{10}
\providecommand{\url}[1]{#1}
\csname url@samestyle\endcsname
\providecommand{\newblock}{\relax}
\providecommand{\bibinfo}[2]{#2}
\providecommand{\BIBentrySTDinterwordspacing}{\spaceskip=0pt\relax}
\providecommand{\BIBentryALTinterwordstretchfactor}{4}
\providecommand{\BIBentryALTinterwordspacing}{\spaceskip=\fontdimen2\font plus
\BIBentryALTinterwordstretchfactor\fontdimen3\font minus
  \fontdimen4\font\relax}
\providecommand{\BIBforeignlanguage}[2]{{%
\expandafter\ifx\csname l@#1\endcsname\relax
\typeout{** WARNING: IEEEtran.bst: No hyphenation pattern has been}%
\typeout{** loaded for the language `#1'. Using the pattern for}%
\typeout{** the default language instead.}%
\else
\language=\csname l@#1\endcsname
\fi
#2}}
\providecommand{\BIBdecl}{\relax}
\BIBdecl

\bibitem{nvidia}
M.~Bojarski, D.~D. Testa, D.~Dworakowski, B.~Firner, B.~Flepp, P.~Goyal, L.~D.
  Jackel, M.~Monfort, U.~Muller, J.~Zhang, X.~Zhang, J.~Zhao, and K.~Zieba,
  ``End to end learning for self-driving cars,'' 2016.

\bibitem{GLU12}
A.~Geiger, P.~Lenz, and R.~Urtasun, ``Are we ready for autonomous driving? the
  kitti vision benchmark suite,'' in \emph{Conference on Computer Vision and
  Pattern Recognition (CVPR)}, 2012.

\bibitem{KYB19}
\BIBentryALTinterwordspacing
Y.~Kang, H.~Yin, and C.~Berger, ``{Test your self-driving algorithm: An
  overview of publicly available driving datasets and virtual testing
  environments},'' \emph{IEEE Transactions on Intelligent Vehicles}, vol.~4,
  no.~2, pp. 171--185, Mar. 2019. [Online]. Available:
  \url{https://ieeexplore.ieee.org/document/8667012}
\BIBentrySTDinterwordspacing

\bibitem{KC18}
\BIBentryALTinterwordspacing
M.~Koziarski and B.~Cyganek, ``Impact of low resolution on image recognition
  with deep neural networks: An experimental study,'' \emph{International
  Journal of Applied Mathematics and Computer Science}, vol.~28, no.~4, pp.
  735--744, 2018. [Online]. Available:
  \url{https://content.sciendo.com/view/journals/amcs/28/4/article-p735.xml}
\BIBentrySTDinterwordspacing

\bibitem{HXP17}
H.~Hosseini, B.~Xiao, and R.~Poovendran, ``Google's cloud vision api is not
  robust to noise,'' Apr. 2017.

\bibitem{SZ14}
K.~Simonyan and A.~Zisserman, ``Very deep convolutional networks for
  large-scale image recognition,'' Sep. 2014.

\bibitem{BVPB19}
A.~Bianchi, M.~R. Vendra, P.~Protopapas, and M.~Brambilla, ``Improving image
  classification robustness through selective cnn-filters fine-tuning,'' Mar.
  2019.

\bibitem{SOZ+18}
Z.~{Sun}, M.~{Ozay}, Y.~{Zhang}, X.~{Liu}, and T.~{Okatani}, ``Feature
  quantization for defending against distortion of images,'' in \emph{2018
  IEEE/CVF Conference on Computer Vision and Pattern Recognition}, 2018, pp.
  7957--7966.

\bibitem{RGBP18}
P.~Roy, S.~Ghosh, S.~Bhattacharya, and U.~Pal, ``Effects of degradations on
  deep neural network architectures,'' Jul. 2019.

\bibitem{GSA+18}
S.~{Ghosh}, R.~{Shet}, P.~{Amon}, A.~{Hutter}, and A.~{Kaup}, ``Robustness of
  deep convolutional neural networks for image degradations,'' in \emph{2018
  IEEE International Conference on Acoustics, Speech and Signal Processing
  (ICASSP)}, 2018, pp. 2916--2920.

\bibitem{UC18}
M.~Usama and D.~E. Chang, ``Towards robust neural networks with lipschitz
  continuity,'' Nov. 2018.

\bibitem{RDGF16}
J.~Redmon, S.~Divvala, R.~Girshick, and A.~Farhadi, ``You only look once:
  Unified, real-time object detection,'' 2016.

\end{thebibliography}

\end{document}